\documentclass[lettersize,journal]{IEEEtran}
\usepackage{amsmath,amsfonts}
\usepackage{algorithmic}
\usepackage{algorithm}

\usepackage{array}
\usepackage[caption=false,font=normalsize,labelfont=sf,textfont=sf]{subfig}
\usepackage{textcomp}
\usepackage{stfloats}
\usepackage{url}
\usepackage{verbatim}
\usepackage{graphicx}
\usepackage{booktabs}
\usepackage{cite}
\usepackage{amssymb}
\usepackage{amsbsy}
\usepackage{bm}
\usepackage{caption}
\usepackage{tabularx}
\usepackage{nomencl}
\usepackage{cancel}
\usepackage{graphicx}
\usepackage{subcaption}
\usepackage{multirow}
\usepackage{tikz}
\usepackage{pgfplots}
\usepackage{arydshln}
\usepackage{soul}

\definecolor{darkergreen}{rgb}{0.0, 0.5, 0.0}
\definecolor{dgreen}{rgb}{0.3, 0.8, 0.0}

\definecolor{darkeryellow}{rgb}{0.8, 0.8, 0.0} 
\definecolor{darkerorange}{rgb}{1.0, 0.6, 0.0}
\definecolor{darkerred}{rgb}{2.0, 0.0, 0.0}

\usepackage{threeparttable}

\makenomenclature

\renewcommand{\nompreamble}{The below list presents several symbols that will be used within the body of the paper.}

\usepackage{etoolbox}
\renewcommand\nomgroup[1]{%
  \item[\bfseries
  \ifstrequal{#1}{M}{HDRM Parameters}{%
  \ifstrequal{#1}{O}{Robust Subsystem-Based Adaptive Control Parameters}{%
  \ifstrequal{#1}{E}{Electromechanical Linear Actuator Parameters}{%
  \ifstrequal{#1}{O}{Other symbols}{}}}}%
]}

\usepackage{siunitx}
\usepackage{amsmath,amssymb,amsfonts}
\usepackage{caption} 
\usepackage{array} 
\usepackage{xcolor} 
\usepackage{setspace} 
\begin{document}

\title{\LARGE \bf{LiDAR-Inertial SLAM-Based Navigation and Safety-Oriented AI-Driven Control System for Skid-Steer Robots}}

\author{Mehdi Heydari Shahna, Eemil Haaparanta, Pauli Mustalahti, and Jouni Mattila
\thanks{Funding for this research was provided by the Business Finland partnership project Future All-Electric Rough Terrain Autonomous Mobile Manipulators (Grant No. 2334/31/2022).}
\thanks{M. H. Shahna, E. Haaparanta, P. Mustalahti, and J. Mattila are with the Faculty of Engineering and Natural Sciences, Tampere University, Finland. \\(e-mail: firstname.lastname@tuni.fi).
        }%
}

\maketitle

\begin{abstract}
Integrating artificial intelligence (AI) and stochastic technologies into the mobile robot navigation and control (MRNC) framework while adhering to rigorous safety standards presents significant challenges. To address these challenges, this paper proposes a comprehensively integrated MRNC framework for skid-steer wheeled mobile robots (SSWMRs), in which all components are actively engaged in real-time execution. The framework comprises: 1) a LiDAR-inertial simultaneous localization and mapping (SLAM) algorithm for estimating the current pose of the robot within the built map; 2) an effective path-following control system for generating desired linear and angular velocity commands based on the current pose and the desired pose; 3) inverse kinematics for transferring linear and angular velocity commands into left and right side velocity commands; and 4) a robust AI-driven (RAID) control system incorporating a radial basis function network (RBFN) with a new adaptive algorithm to enforce in-wheel actuation systems to track each side motion commands.
To further meet safety requirements, the proposed RAID control within the MRNC framework of the SSWMR constrains AI-generated tracking performance within predefined overshoot and steady-state error limits, while ensuring robustness and system stability by compensating for modeling errors, unknown RBF weights, and external forces.
Experimental results verify the proposed MRNC framework performance for a $\bm{4,836}$ kg SSWMR operating on soft terrain.
\end{abstract}


\section{Introduction}
Skid-steer wheeled robots (SSWMRs) are a common choice for tasks performed in off-road settings due to their straightforward design, ability to turn in place, and high maneuverability. However, they exhibit complex wheel-ground interactions and require an advanced integration of technologies within the mobile robot navigation and control (MRNC) framework to operate autonomously under uncertainties and disturbances. Hence, ensuring compatibility in integration and control while adhering to high safety standards is a significant challenge \cite{khan2021comprehensive}. MRNC involves executing a series of actions that move it from its current position to the intended destination. This requires the robot to accurately determine its current location. Simultaneous localization and mapping (SLAM) enables a robot to concurrently build a map of an unknown environment while estimating its pose. Most SLAM methods are based on either visual or LiDAR technology.\cite{labbe2019rtab}. Although visual SLAM is a more affordable option that relies on camera images, performing well in feature-rich environments with good lighting, it struggles in low-light or feature-poor conditions. In addition, it is prone to drift, particularly over extended operation \cite{mohamed2019survey, cadena2016past}. On the other hand, LiDAR-based SLAM uses laser scanning data to provide highly accurate depth information, making it more robust in diverse conditions, including low-light and feature-poor areas. A state-of-the-art approach to improve pose estimation in SLAM is sensor fusion, where data from multiple sensors are integrated to enhance robustness, especially beneficial in environments with sensory degradation caused by factors such as geometric ambiguities, illumination variations, or environmental obscurants (e.g., dust, smoke, fog, rain, or snowfall) \cite{khattak2020complementary, shan2020lio}. To obtain more accurate positioning and create better maps, LiDAR data can be combined with information from other sensors like GPS and IMUs. Real-time SSWMR trajectory estimation is accomplished in this paper through the use of the LIO-SAM algorithm, a technique derived from the findings of \cite{shan2020lio} that performs LiDAR inertial odometry via smoothing and mapping.

\begin{figure}[h] 
  \centering
\scalebox{1}
    {\includegraphics[trim={0cm 0.0cm 0.0cm
    0cm},clip,width=\columnwidth]{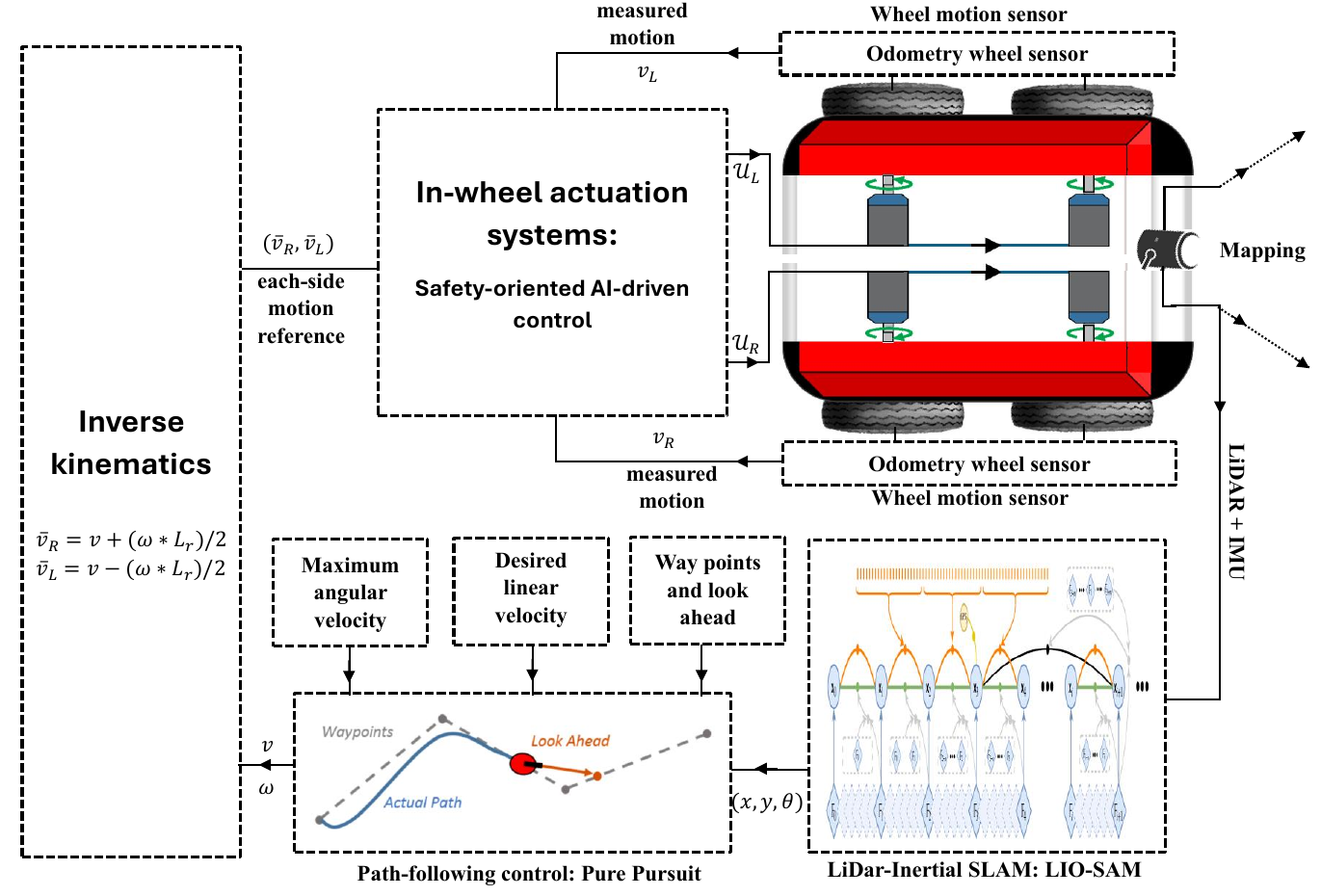}}
  \caption{The whole proposed MRNC framework for SSWMRs.}
  \label{schematicwhole}
\end{figure}

With this essential information about the SSWMR pose in a built map, the control system must then develop a robust tracking strategy to reach the desired motion by continuously generating appropriate control signals \cite{huskic2019gerona}, monitoring motion feedback, and compensating for external disturbances such as uneven terrain, load variations, or environmental forces \cite{shahna2025fault}. The use of artificial intelligence (AI) in state-of-the-art control system technologies has grown rapidly due to benefits such as adaptability to different environments and autonomy and learning decision-making for reducing the need for human oversight, thereby lowering operational costs while increasing productivity \cite{lee2024learning, shahna2024integrating}. However, using AI-driven control technologies in the MRNC framework while adhering to rigorous safety standards presents significant challenges \cite{carabantes2020black, ISO81283}. According to ISO/IEC TR 5469 \cite{ISO81283}, a supervisory control module using methods like barrier function approaches can ensure the system remains within a safe operating region. Hence, to avoid violations of safety-related system
constraints, \cite{shi2024barrier} employed a logarithmic barrier Lyapunov function (BLF) through an adaptive neural network (NN) optimized control for marine vessels suffering from ocean disturbances. However, it assumed a constant state constraint that does not directly guarantee restrictions on control performance metrics such as overshooting or steady-state error. Although \cite{han2013barrier} proposed a fractional BLF through sliding-mode adaptive radial basis function networks (RBFNs) with predefined control performance, the study overlooked the safety-defined constraints on input control signals, which are limited by mechanical restrictions in the actuator's working space.
Building on \cite{shahna2025model, shahna2024robusttttsdf} and taking a step further in meeting safety requirements, this paper proposes a comprehensive control system within the MRNC framework of SSWMR to constrain AI-generated tracking performance within predefined overshoot and steady-state error boundaries as well as safety-defined control inputs. The control system design includes: 1) A pure pursuit algorithm forpath-following control for generating desired linear and angular velocity commands of the robot based on the current pose (generated by LIO-SLAM) and the desired pose; 2) inverse kinematics for transferring linear and angular velocity commands into left- and right-side linear velocity commands; and 3) a robust AI-driven (RAID) control system incorporating RBFNs with a new adaptive algorithm to enforce in-wheel actuation systems to track each side motion commands within the predefined performance.
The contributions of the study include:
1) this is the first paper to present a comprehensive MRNC framework for SSWMRs, integrating a state-of-the-art LiDAR-inertial SLAM-based navigation with a novel model-based, safety-oriented, AI-driven control system; 2) unlike \cite{zhang2019low, han2013barrier, shi2024barrier}, safety-defined input-output constraints are addressed in the design of the RAID control by using enhanced logarithmic BLF with user-defined transient and steady-state performance;
3) employing the RAID control guarantees that the actual motion of SSWMRs tracks the reference motion with an exponential convergence rate, even under modeling errors and external disturbances; and 4) the proposed MRNC performance is validated by implementing it on a $4,836$ kg SSWMR operating on a slippery and wet terrain during low-light conditions. The entire proposed MRNC framework for SSWMRs is shown in Fig. \ref{schematicwhole}, where the integrated framework and the displayed parameters will be elaborated on later.

The remainder of this paper is organized as follows: Section II introduces the LiDAR-inertial SLAM algorithm used for SSWMRs, based on the LIO-SAM algorithm, which lays the foundation for the control system. Section III designs the proposed control system, including path-following control, inverse kinematics, and the safety-oriented AI-driven in-wheel actuation system control. In addition, this section provides a rigorous stability proof. In Section IV, the proposed MRNC framework is experimentally applied to a heavy-weight SSWMR to demonstrate the MRNC framework's effectiveness.

\section{Lidar-Inertial Slam-Based Navigation}
\begin{figure}[h] 
  \centering
\scalebox{1}
    {\includegraphics[trim={0cm 0.0cm 0.0cm
    0cm},clip,width=\columnwidth]{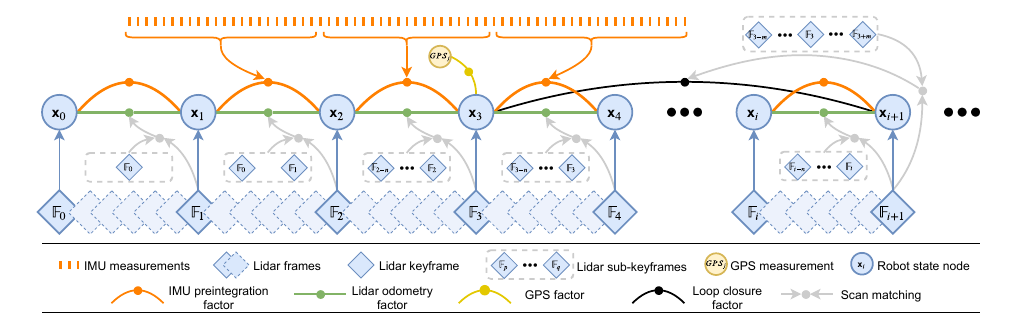}}
  \caption{The system structure of LIO-SAM \cite{shan2020lio}.}
  \label{liuooo}
\end{figure}

Building upon the work of \cite{shan2020lio}, this paper employs the LIO-SAM algorithm. This system uses a factor graph to combine data from LiDAR and inertial sensors for accurate motion tracking. It integrates various types of measurements, including loop closures identified through radius-search, into this graph. The IMU data is processed to correct LiDAR point cloud distortion and provide an initial motion estimate. This estimate then helps refine the IMU's bias. To achieve real-time speed, the system optimizes poses using only recent LiDAR scans, rather than a full global map. This local scan-matching approach, along with strategic keyframe selection and a sliding window technique, significantly improves efficiency. Specifically, new keyframes are compared to a limited set of recent, representative poses (sub-keyframes).
Fig. \ref{liuooo} illustrates the factor graph architecture that forms the SLAM system basis. It integrates data from an IMU, a LiDAR, and, when available, a GPS. To build this graph, four distinct factor types are incorporated: (a) factors derived from IMU preintegration, (b) LiDAR odometry factors, (c) GPS measurement factors, and (d) loop closure factors, as detailed in \cite{shan2020lio}.

\vspace{-0.6cm}

\subsection{Lidar Odometry}\label{section1}
The LiDAR odometry pipeline of LIO-SAM can be separated into distinct parts: feature extraction, voxel map creation, correspondence matching, and transform optimization. These modules follow the LiDAR odometry algorithm presented in \cite{zhang2017low} and \cite{shan2020lio}. Features are extracted from each incoming LiDAR scan. The features are divided into edge and planar features, depending on the smoothness value $c$ of a local surface defined as:
\begin{equation}
\small
\begin{aligned}
\small
\label{equation:adjusted_velocities}
c=\frac{1}{|\mathcal{S}| \cdot \| p_{i+1, k}| |}\left\|\sum_{u \in \mathcal{S}, u \neq k}\left(p_{i+1, k}-p_{i+1, u}\right)\right\|
\end{aligned}
\end{equation}

where $p_{i+1}$ is a point represented in the LiDAR frame at index $k$ from the current LiDAR scan at time $t=i+1$. The set $\mathcal{S}$ contains consecutive points distributed equally around point $k$. The normalized sum of differences between the point $k$ and points $u \in \mathcal{S}$ classifies if a point is considered as a feature. Points with the largest value of $c$ are selected as edge points, as this would indicate a greater deviation between local points. Accordingly, points attaining the smallest smoothness values are considered as planar points \cite{zhang2017low}.
Every LiDAR frame $\mathbb{F}_i$ is composed of the extracted planar features $F_i^p$ and edge features $F_i^e$ represented in a body-fixed frame. For efficient processing, LiDAR frame $\mathbb{F}_{i+1}$ becomes a keyframe only if the robot's movement significantly differs from its prior position, as determined by a set threshold. This frame is linked to the new node in the factor graph \cite{grisetti2010tutorial}.
A point cloud map is created using the $n$ most recent keyframes, called sub-keyframes. Employing the associated transformations $\left\{\mathbf{T}_{i-n}, \ldots, \mathbf{T}_i\right\}$, the sub-keyframes $\left\{\mathbb{F}_{i-n}, \ldots, \mathbb{F}_i\right\}$ become transformed to a fixed world frame. The point cloud map is represented as a voxel map $\mathbf{M}_i$ comprising sub-voxel maps $\mathbf{M}_i^p$ and $\mathbf{M}_i^e$ for transformed planar and edge features, respectively. The subvoxel maps are defined as the intersection of the transformed feature points to avoid duplicate features inside single voxels:

\begin{equation}
\begin{aligned}
\small
\label{equation:adjusted_velocities}
\begin{gathered}
\mathbf{M}_i^p={ }^{\prime} F_i^p \cap \cap^{\prime} F_{i-1}^p \cap \ldots \cap^{\prime} F_{i-n}^p \\
\mathbf{M}_i^e={ }^{\prime} F_i^e \cap \cap^{\prime} F_{i-1}^e \cap \ldots \cap^{\prime} F_{i-n}^e,
\end{gathered}
\end{aligned}
\end{equation}
where ${ }^{\prime}F_i^p$ and ${ }^{\prime}F_i^e$ denote the point sets transformed from the body-fixed frame to world-fixed one. The initial transform is given by an IMU measurement \cite{shan2020lio}.
A kD-tree is built for both local planar and surface maps. This allows the use of a nearest neighbor search to find the closest points for the extracted feature points ${ }^{\prime}F_{i+1}^p$ and ${ }^{\prime} F_{i+1}^e$ from the current scan. The neighboring points are used for distance calculation between a feature point and its correspondent edge line in $\mathbf{M}_i^e$ and between a feature point and its correspondent planar patch in $\mathbf{M}_i^p$. The edge-to-line distance is defined as

\begin{equation}
\begin{aligned}
\small
\label{equation:adjusted_velocities}
d_{e_k}=\frac{\left|\left(p_{i+1, k}^e-p_{i, u}^e\right) \times\left(p_{i+1, k}^e-p_{i, v}^e\right)\right|}{\left|p_{i, u}^e-p_{i, v}^e\right|}
\end{aligned}
\end{equation}

The points $p_{i, u}^e$ and $p_{i, v}^e$ form an edge line in $\mathbf{M}_i^e$. The distance metric for point-to-plane is:

\begin{equation}
\begin{aligned}
\small
\label{equation:adjusted_velocities}
d_{p_k}=\frac{\left|\begin{array}{c}
\left(p_{i+1, k}^p-p_{i, u}^p\right) \\
\left(p_{i, u}^p-p_{i, v}^p\right) \times\left(p_{i, u}^p-p_{i, w}^p\right)
\end{array}\right|}{\left|\left(p_{i, u}^p-p_{i, v}^p\right) \times\left(p_{i, u}^p-p_{i, w}^p\right)\right|}
\end{aligned}
\end{equation}

The points $p_{i, u}^p, p_{i, v}^p$ and $p_{i, w}^p$ form the correspondent planar patch in $\mathbf{M}_i^p$ \cite{zhang2017low}. Using the functions above, the optimal transform can be solved by minimizing point-to-plane and point-to-edge distances using the iterative Gauss-Newton optimization:

\begin{equation}
\begin{aligned}
\small
\label{equation:adjusted_velocities}
\min _{\mathbf{T}_{i+1}}\left\{\sum_{p_{i+1, k}^e \in^{\prime} F_{i+1}^e} d_{e_k}+\sum_{p_{i+1, k}^p \in^{\prime} F_{i+1}^p} d_{p_k}\right\}
\end{aligned}
\end{equation}
The LiDAR odometry factor, which is the relative transformation between poses at $i$ and ${i+1}$, is then obtained by \cite{shan2020lio} $\Delta \mathbf{T}_{i, i+1}=\mathbf{T}_i^T \mathbf{T}_{i+1}$.

\subsection{Loop closure: radius search}\label{section2}
To address drifts, the default loop closure presented in \cite{shan2020lio} is a Euclidean distance-based solution referred to as radius search. To detect loop closures, the system checks prior states located within a radius around each newly added state in the factor graph. The transformation $\Delta T_{n, i+1}$ between the current frame $\mathbb{F}_{i+1}$ and the prior sub-keyframes $\left\{\mathbb{F}_{n-m}, \ldots \mathbb{F}_n, \ldots, \mathbb{F}_{n+m}\right\}$ is obtained through the scan matching procedure described in the LiDAR odometry section. The transformation obtained through scan matching is added as a loop closure factor to the factor graph. $\mathbb{F}_n$ is the loop closure candidate frame and $m$ is a sliding window parameter that determines the number of sub-keyframes chosen for scan matching.
In summary,  LiDAR provides spatial measurements for mapping, while the IMU refines motion estimation, especially in featureless areas. The system outputs a consistent map of the environment and the robot's pose estimation $(x, y, \theta)$, representing the position and orientation in a 2D plane.

\section{Design of Control System}
\subsection{Path-Following Control}
\label{puress}
The pure pursuit algorithm is a straightforward method for implementation on SSWMRs \cite{krecht2020possible}. It designates a target point ahead of the robot based on the current pose \cite{krecht2020possible, coulter1992implementation} and conveys it to the in-wheel actuation system for control purposes. The next target point (waypoint) \((x_t, y_t)\) is selected at the look-ahead distance \(L\) from the path such that the distance from the robot's current position \((x, y)\) is at least $D_t \geq L > 0$.
Then, the angle between the robot's current position and the target point is obtained as  \(\alpha_t=\text{atan2}(y_t - y, x_t - x)\). The steering angle can be calculated as
$\delta_t = \text{atan2}( \frac{2L \sin(\alpha_t)}{D_t})$.
Finally, the commanded angular velocity based on the desired linear velocity $v$ for the next step time can be obtained as:
$\omega = \frac{v}{\frac{D_t}{2 \tan(\delta_t)}}$.

\textit{Remark 1}: atan2 $\left(.\right)$ function is inherently designed to handle singularities gracefully and returns the four-quadrant inverse tangent (tan$^{-1}$), which must be real in the closed interval $(–\pi, \pi)$. In addition, if \( \delta_t = 0 \), then \( \omega \) becomes infinite. To avoid singularity, we set \( \omega = 0 \) if $\delta = 0$. Finally, after reaching the last waypoint, both the angular velocity $\omega$ and linear velocity $v$ are set to zero to ensure the robot stops at the last waypoint without any further velocity commands.

\begin{figure}[h] 
  \centering
\scalebox{0.75}
    {\includegraphics[trim={0cm 0.0cm 0.0cm
    0cm},clip,width=\columnwidth]{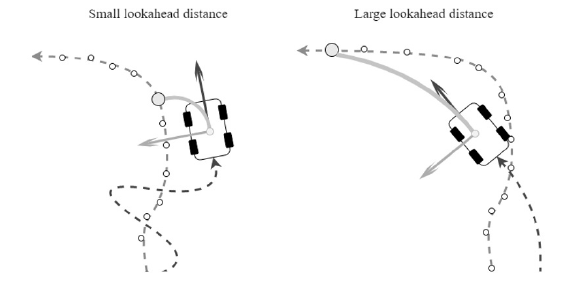}}
  \caption{Impacts of lookahead distance \cite{krecht2020possible}.}
  \label{persuit}
\end{figure}

The algorithm's successful application depends on the proper size of the lookahead distance \cite{krecht2020possible}. As shown in Fig. \ref{persuit}, if the lookahead distance is too small, oscillation might occur. In contrast, in case of an inconveniently large lookahead distance, the problem of truncating might be risen.

\subsection{Inverse Kinematics of SSWMRs}
\label{inverce}
The inverse kinematics of SSWMRs from the output of Section \ref{puress} can be provided as:
\begin{equation}
\begin{aligned}
\small
\label{equataadadaion:adjusted_velocities}
& \bar{v}_R=v+(\omega  L_r) / 2, \hspace{0.2cm} \bar{v}_L=v-(\omega L_r) / 2
\end{aligned}
\end{equation}
$L_r$ represents the wheelbase width of the robot. Note, based on Section \ref{puress} and Eq. \eqref{equataadadaion:adjusted_velocities}, the desired linear velocity robot must be set as:
$v\leq \frac{v_{max}}{2(1 + \tan(\delta_t) \frac{L_r}{D_t})}$,
where $v_{max}$ is the predefined maximum velocity of the SSWMR. These outputs $\left(\bar{v}_R, \bar{v}_L\right)$ are then sent to both sides of in-wheel actuation controllers as references, ensuring proper execution of the motion.

\subsection{Safety-Oriented RAID Control for In-wheel Actuators}
We define the linear velocity for each side ($i: R, L$) as $v_i \in \mathbb{R}$. The motion dynamics of the in-wheel actuation system is
\begin{equation}
\begin{aligned}
\small
\label{3asd3}
\dot{v}_i =& g_{i}\left(v_i, t\right) \mathcal{U}_i (t) + d_i\left(v_i, t\right)+\Delta_{i}(t)
\end{aligned}
\end{equation}
where $d_i: \mathbb{R} \times \mathbb{R}_{+} \rightarrow \mathbb{R}$ represents the bounded modeling uncertainties, $\mathcal{U}_i:\mathbb{R}_{+} \rightarrow \mathbb{R}$ denotes the model-based control generated by the actuator mechanism, and $\Delta_i: \mathbb{R}_{+} \rightarrow \mathbb{R}$ accounts for the bounded external disturbance. Note that $g_i: \mathbb{R} \times \mathbb{R}_{+} \rightarrow \mathbb{R}_{+}$ is the control coefficient.

\textit{Assumption 1}: $g_i$ is allowed to be non-constant and unknown, but it must be positive, ensuring that the control actuator signal has the same sign as the robot motion.

\textit{Assumption 2} \cite{huang2019practical, shahna2025model}: Assume positive constants $g_i^* \in \mathbb{R}_{+}$ and $\Delta_i^* \in \mathbb{R}_{+}$ for bounds of functional control gain and external disturbances as $|g_i(v_i, t)| \leq g_i^*$ and $|\Delta_i (t)| \leq \Delta_i^*$.

To define safety-oriented control inputs for preventing harm to mechanical actuators, let us define the following constraints: 
\begin{equation}
\begin{aligned}
\small
\label{33}
\lambda_{i}= \begin{cases}\frac{1}{\left|\mathcal{U}_i(t)\right|+1}, & \mathcal{U}_i(t) \geq \bar{\mathcal{U}}_i \text { or } \mathcal{U}_i(t) \leq \underline{\mathcal{U}}_i \\ 1 & \underline{\mathcal{U}}_i \leq \mathcal{U}_i(t) \leq \bar{\mathcal{U}}_i\end{cases}
\end{aligned}
\end{equation}
and
\begin{equation}
\begin{aligned}
\small
\label{33}
\bar{\lambda}_{i}= \begin{cases}\bar{\mathcal{U}}_i-\frac{\mathcal{U}_i(t)}{\mid \mathcal{U}_i(t)+1}, & \mathcal{U}_i(t) \geq \bar{\mathcal{U}}_i \\ 0 & \underline{\mathcal{U}}_i \leq \mathcal{U}_i(t) \leq \bar{\mathcal{U}}_i \\ \underline{\mathcal{U}}_i-\frac{\mathcal{U}_i(t)}{\left|\mathcal{U}_i(t)\right|+1} & \mathcal{U}_i(t) \leq \underline{\mathcal{U}}_i\end{cases}
\end{aligned}
\end{equation}
$\underline{\mathcal{U}}_i$ and $\bar{\mathcal{U}}_i$ are the acceptable boundaries following mechanical actuator specifications, respectively.
Then, if we define safety-constrained control inputs as
\begin{equation}
\begin{aligned}
\small
\label{33asdad}
\mathcal{U}^*_i=&\lambda_i \mathcal{U}_i+\bar{\lambda}_i
\end{aligned}
\end{equation}
we will always have $\mathcal{U}^*_i$, as
\begin{equation}
\begin{aligned}
\small
\label{asdasd}
\mathcal{U}^*_i= \begin{cases}\bar{\mathcal{U}}_i, & \mathcal{U}_i(t) \geq \bar{\mathcal{U}}_i \\ {\mathcal{U}}_i  & \underline{\mathcal{U}}_i \leq \mathcal{U}_i(t) \leq \bar{\mathcal{U}}_i \\ \underline{\mathcal{U}}_i & \mathcal{U}_i(t) \leq \underline{\mathcal{U}}_i\end{cases}
\end{aligned}
\end{equation}
We define the tracking error as $e_i= v_i - \bar{v}_{i}$, where $\bar{v}_i$ is the reference velocity of $i$ side received from Section \ref{inverce}. By altering control inputs $\mathcal{U}_i$ with $\mathcal{U}^*_i$ in \eqref{3asd3}, we have
\begin{equation}
\begin{aligned}
\small
\label{3ghxhnfg3}
\dot{e}_i =& g_{i} \mathcal{U}^*_i+ d_i+\Delta_{i}-\dot{\bar{v}}_i
\end{aligned}
\end{equation}
By employing RBFNs, the uncertainty term can be defined as
\begin{equation}
\begin{aligned}
\small
\label{asdfrg}
d_i\left(v_i\right)=\left(a_i\right)^{\top} \Phi_i\left(v_i\right)+\delta_i\left(v_i\right)
\end{aligned}
\end{equation}
where $v_i$ serves as the input to the RBFNs. The unknown output layer's weight vector $a_i \in  \mathbb{R}^{N_i}$ corresponds to $N_i$ hidden-layer neurons. $\delta_i(.): \mathbb{R} \rightarrow \mathbb{R}$ represents the unknown approximation error. $\Phi(.): \mathbb{R} \rightarrow R^{N_i}$ is the function defined as
\begin{equation}
\begin{aligned}
\small
\label{e5}
\Phi_i\left(v_i\right)=\left[\begin{array}{lll}
\Phi_{i 1}\left(v_i\right) & \Phi_{i 2}\left(v_i\right) \cdots \Phi_{i N_i}\left(v_i\right)
\end{array}\right]^{\top}
\end{aligned}
\end{equation}
$\Phi_{i k}\left(.\right)$ can be selected a common Gaussian functions, as
\begin{equation}
\begin{aligned}
\small
\label{equation:adjusted_velocities}
\Phi_{i k}\left(v_i\right)=\exp \left(-\frac{\left\|v_i-{\alpha}_{i k}\right\|^2}{\gamma_{i k}^2}\right), \quad k=1,2, \ldots, N_i
\end{aligned}
\end{equation}

where $\alpha_{i k} \in \mathbb{R}$ can be selected stochastically and defines the center of the receptive field, while $\gamma_{i k} \in \mathbb{R} $ is a given constant determining the width of the Gaussian functions. The symbol $\|.\|$ denotes the Euclidean norm. 

\textit{Assumption 3}: Suppose the unknown approximation error satisfies $\left|\delta_i\left(v_i\right)\right| \leq {\delta}^*_i<\infty$. However, the weights $a_i$, the approximation error $\delta_i$, and its upper bound ${\delta}^*_i$ are unknown.

From \eqref{3ghxhnfg3} and \eqref{asdfrg}, we can have:
\begin{equation}
\begin{aligned}
\small
\label{33asdas}
\dot{e}_i =& g_{i} \mathcal{U}^*_i+ \left(a_i\right)^{\top} \Phi_i\left(v_i\right)+\delta_i\left(v_i\right)+\Delta_{i}-\dot{\bar{v}}_i
\end{aligned}
\end{equation}

According to ISO/IEC TR 5469 \cite{ISO81283}, safety-defined constraints can be formulated as barrier functions to define an invariant set. This method ensures that the system remains within a safe operating region. Let us define the bound of tracking control performance in terms of overshooting and steady-state error as follows $ |{e}_{i}| < o_i(t)$, where:
\begin{equation}
\begin{aligned}
\small
\label{20}
o_i=\left(o^{ov}_i-o^{b}_i\right) e^{-o_i^* t}+o^{b}_i
\end{aligned}
\end{equation}
where $o^{ov}_i$,$o_i^{b},$ and $o_i^*$ are positive constants. The transient response of $e_i$ must be restricted to $o_i^{o v}$. The exponential convergence rate and steady state response of $e_i$ are $o_i^*$ and $o_i^b$. The model-based constrained control input in Eq. \eqref{33asdad} is defined as
\begin{equation}
\begin{aligned}
\small
\label{2fdgdffg0}
\mathcal{U}^*_i=\lambda_i \mathcal{U}_i+\bar{\lambda}_i=\lambda_i [u_{i}+ f_i \left(v_i\right)]+\bar{\lambda}_i
\end{aligned}
\end{equation}
where $f_i: \mathbb{R} \times \mathbb{R}_{+} \rightarrow \mathbb{R}$ is the known modeling term of the in-wheel actuation system. To constrain system states to be within the safety-defined tracking performance, we can use singularities for designing the control term $u_i$ in which by choosing $|{e}_i(t_0)|<o_i(t_0)$, the control input must tend toward infinity as ${e}_i$ approaches $o_i$. In this case, an error such as ``Infinity or NaN value encountered" will occur in programming environments, causing the execution to halt and triggering emergency safety stops. Let us propose a novel RAID control $u_i$, as
\begin{equation}
\begin{aligned}
\small
\label{aghashgsfz}
u_{i}=&- f_i-\frac{1}{2} \beta_i e_i-\frac{{\dot{o}_i}^2 e^3_i}{o^2_i(o^2_i - e^2_i)}\\
&- \frac{{e}_i}{o^2_i - e^2_i} (\hat{\phi}^4_i \|\Phi_i\|+\|\Phi_i\|^2+\dot{\bar{v}}_i^2+1)
\end{aligned}
\end{equation}
where $\beta_i>0$ is the control design parameter. $\hat{\phi}_i$ is the adaptive parameter designed to adjust the weights of the RBFNs in Eq. \eqref{asdfrg} and compensate for external disturbances in Eq. \eqref{33asdas}. The adaptive law is defined as
\begin{equation}
\begin{aligned}
\small
\label{fshbdgjhs}
\dot{\hat{\phi}}_i=&-\frac{1}{2}\kappa_i \hat{\phi}_i+\left(\frac{{e}_i}{o^2_i - e^2_i}\right)^2 \hat{\phi}^3_i \|\Phi_i\|
\end{aligned}
\end{equation}
where $\kappa_i>0$. Eqs. \eqref{aghashgsfz} and \eqref{fshbdgjhs} contain the fractional barrier function with safety-defined singularities.

\indent \textit{Definition 1} \cite{corless1993bounded, shahna2024robustness, heydari2024robusttt} Consider a Lyapunov function $V$ with initial value  $V(t_0)$ related to any system states. For $t \geq t_0$, the system is uniformly exponentially stable if
\begin{equation}
\begin{aligned}
\small
\label{28}
{V} \leq \bar{c} {V}\left(t_0\right) e^{-{\rho}\left({t-t_0}\right)}+ \hspace{0.1cm}{\ell} \hspace{0.1cm} {{\rho}}^{-1}
\end{aligned}
\end{equation}
where $\bar{c}$, $\rho$, and $\ell$ are positive constants.

\textit{Theorem 1}: Based on \textit{Definition 1},
employing Eqs. \eqref{aghashgsfz} and \eqref{fshbdgjhs} ensures uniformly exponential stability of the system provided in Eq. \eqref{3asd3} in the presence of uncertainties, external disturbance, and predefined safety-related constraints.

\textit{Proof.} See Section \ref{prrof}.

\textit{Lemma 1.} For any positive value of $o_i$, the next inequality is valid for every ${e}_i$ that meets the condition $\left|{e}_i\right|<o_i$ \cite{ren2010adaptive, shahna2025model}:
\begin{equation}
\begin{aligned}
\small
\label{21}
\log \left(\frac{o_i^2}{o^2_i - e^2_i}\right)<\frac{{e}_i^2}{o^2_i - e^2_i}
\end{aligned}
\end{equation}

\textit{Remark 2}: Based on the Cauchy-Schwarz inequality for any $a$ and $b$ we can say $ab \leq c a^2 + \frac{1}{4c} b^2$, where $c$ is positive.

\subsection{Stability Analysis}
\label{prrof}
Let us introduce a BLF, as
\begin{equation}
\begin{aligned}
\small
\label{97asdsada}
{V}_i=\frac{1}{2} \log \left(\frac{o_i^2}{o^2_i - e^2_i}\right)+\frac{1}{2} \lambda_{{min}_i} \hat{\phi}_i^2
\end{aligned}
\end{equation}
where $\lambda_{{min}_i}$ is an unknown constant. After differentiating Eq. \eqref{97asdsada}, we have
\begin{equation}
\begin{aligned}
\small
\dot{V}_i=\frac{-\dot{o}_i e_i^2 +e_i o_i \dot{e}_i}{o_i(o^2_i - e^2_i)} + \lambda_{{min}_i} \hat{\phi}_i \dot{\hat{\phi}}_i
\end{aligned}
\end{equation}
By inserting Eq. \eqref{33asdas}, we have
\begin{equation}
\begin{aligned}
\small
\dot{V}_i =& \frac{-\dot{o}_i e^2_i}{o_i(o^2_i - e^2_i)}+ \frac{e_i}{o^2_i - e^2_i} (g_{i} \mathcal{U}^*_i + a_i^{\top} \Phi_i+\delta_i\\
&+\Delta_{i}-\dot{\bar{v}}_i)+ \lambda_{{min}_i} \hat{\phi}_i \dot{\hat{\phi}}_i
\end{aligned}
\end{equation}
From \eqref{2fdgdffg0},
\begin{equation}
\begin{aligned}
\small
\dot{V}_i =& \frac{-\dot{o}_i e^2_i}{o_i(o^2_i - e^2_i)}+ \frac{e_i}{o^2_i - e^2_i} (g_{i} \lambda_i u_{i} + g_{i} \lambda_i f_i \\
&+ a_i^{\top} \Phi_i+\delta_i+g_{i}\bar{\lambda}_i+\Delta_{i}-\dot{\bar{v}}_i)+ \lambda_{{min}_i} \hat{\phi}_i \dot{\hat{\phi}}_i
\end{aligned}
\end{equation}
Based on \textit{Remark 2},
\begin{equation}
\begin{aligned}
\small
\dot{V}_i \leq& \lambda_{{min}_i}\frac{\dot{o}^2_i e^4_i}{o^2_i(o^2_i - e^2_i)^2}+\frac{1}{4\lambda_{{min}_i}}+ \frac{e_i}{o^2_i - e^2_i} (g_{i} \lambda_i u_{i}  \\
&+ g_{i} \lambda_i f_i)+\lambda_{{min}_i}\frac{e^2_i}{(o^2_i - e^2_i)^2}\dot{\bar{v}}_i^2+\frac{1}{4\lambda_{{min}_i}}\\
&+\lambda_{{min}_i}\frac{e^2_i}{(o^2_i - e^2_i)^2} \|\Phi_i\|^2+\frac{1}{4 \lambda_{{min}_i}}\|a_i\|^2\\
&+\lambda_{{min}_i}\frac{e^2_i}{(o^2_i - e^2_i)^2}+\frac{1}{4 \lambda_{{min}_i}}(g_{i}\bar{\lambda}_i+\Delta_{i}+ \delta_i)^2\\
&+ \lambda_{{min}_i} \hat{\phi}_i \dot{\hat{\phi}}_i
\end{aligned}
\end{equation}
After mathematical simplifications,
\begin{equation}
\begin{aligned}
\small
\dot{V}_i \leq& \lambda_{{min}_i}\frac{\dot{o}^2_i e^4_i}{o^2_i(o^2_i - e^2_i)^2}+ \frac{e_i}{o^2_i - e^2_i} (g_{i} \lambda_i u_{i}+ g_{i} \lambda_i f_i) \\
&+\lambda_{{min}_i}\frac{e^2_i}{(o^2_i - e^2_i)^2}\dot{\bar{v}}_i^2+\lambda_{{min}_i}\frac{e^2_i}{(o^2_i - e^2_i)^2} \|\Phi_i\|^2\\
&+\lambda_{{min}_i}\frac{e^2_i}{(o^2_i - e^2_i)^2}+ \lambda_{{min}_i} \hat{\phi}_i \dot{\hat{\phi}}_i\\
&+\frac{(g_{i}\bar{\lambda}_i+\Delta_{i}+ \delta_i)^2+\|a_i\|^2+2}{4 \lambda_{{min}_i}}
\end{aligned}
\end{equation}
Inserting the control input $u_i$ and adaptive law from Eqs. \eqref{aghashgsfz} and \eqref{fshbdgjhs}, we obtain
\begin{equation*}
\begin{aligned}
\small
\dot{V}_i \leq& \lambda_{{min}_i}\frac{\dot{o}^2_i e^4_i}{o^2_i(o^2_i - e^2_i)^2}
\end{aligned}
\end{equation*}

\begin{equation}
\begin{aligned}
\small
&-\frac{1}{2}g_{i} \lambda_i \beta_i\frac{e^2_i}{o^2_i - e^2_i} -g_{i} \lambda_i\frac{{\dot{o}_i}^2 e^4_i}{o^2_i(o^2_i - e^2_i)^2}\\
&- g_{i} \lambda_i\frac{{e}^2_i}{(o^2_i - e^2_i)^2} (\hat{\phi}^4_i \|\Phi_i\|+\|\Phi_i\|^2+\dot{\bar{v}}_i^2+1) \\
&+\lambda_{{min}_i}\frac{e^2_i}{(o^2_i - e^2_i)^2}(\dot{\bar{v}}_i^2+ \|\Phi_i\|^2+1+\hat{\phi}^4_i \|\Phi_i\|)\\
&-\frac{1}{2}\lambda_{{min}_i} \kappa_i \hat{\phi}^2_i+\frac{(g_{i}\bar{\lambda}_i+\Delta_{i}+ \delta_i)^2+\|a_i\|^2+2}{4 \lambda_{{min}_i}}
\end{aligned}
\end{equation}
As  $0 < \lambda_{{min}_i} \leq g_i \lambda_i$, based on {Assumptions} 1,2, and 3,
\begin{equation}
\begin{aligned}
\small
\label{sadasdfjksdfv}
\dot{V}_i \leq&-\frac{1}{2} \lambda_{{min}_i} \beta_i\frac{e^2_i}{o^2_i - e^2_i} -\frac{1}{2}\lambda_{{min}_i} \kappa_i \hat{\phi}^2_i\\
&+\frac{(g^*_{i}\bar{\lambda}_i+\Delta^*_{i}+ \delta^*_i)^2+\|a_i\|^2+2}{4 \lambda_{{min}_i}}
\end{aligned}
\end{equation}
Thus, from \textit{Lemma 1}, Eq. \eqref{sadasdfjksdfv} will be
\begin{equation}
\begin{aligned}
\small
\dot{V}_i \leq&-\frac{1}{2} \lambda_{{min}_i} \beta_i\log \left(\frac{o_i^2}{o^2_i - e^2_i}\right)-\frac{1}{2}\lambda_{{min}_i} \kappa_i \hat{\phi}^2_i\\
&+\ell_i
\end{aligned}
\end{equation}
where $\ell_i = \frac{(g^*_{i}\bar{\lambda}_i+\Delta^*_{i}+ \delta^*_i)^2+\|a_i\|^2+2}{4 \lambda_{{min}_i}}$ is a positive constant depending to the intensity of the disturbances, over-saturation control, approximation error, and fitting weights of RBFNs. From \cite{shahna2024robust}, we can say
\begin{equation}
\begin{aligned}
\small
\label{asdas6}
\dot{V}_i \leq -\rho_i V_i+\ell_i, \hspace{0.15cm}
\rho_{i} = \min [\lambda_{{min}_i} \beta_i,\hspace{0.1cm} \kappa_i]
\end{aligned}
\end{equation} 
Now, we define an additional quadratic function based on \eqref{97asdsada} for both sides of the SWMR, as
\begin{equation}
\label{2dfdds2}
\begin{aligned}
\small
V & = V_R+V_L = \frac{1}{2} \sum_{i=R,L} \log \left(\frac{o_i^2}{o^2_i - e^2_i}\right)+ \hat{\phi}_i^2
\end{aligned}
\end{equation} 
By differentiating Eq. \eqref{2dfdds2}, and from \eqref{asdas6}, we obtain
\begin{equation}
\begin{aligned}
\small
\label{6}
\dot{V}=&\dot{V}_R + \dot{V}_L \\
&\leq-\rho_R V_R+\ell_R-\rho_L V_L+\ell_L \leq -\rho V+\ell
\end{aligned}
\end{equation}
where
\begin{equation}
\begin{aligned}
\small
\label{44}
\rho &= \min [\rho_R,\hspace{0.1cm} \rho_L], \hspace{0.2cm} \ell=\ell_R+\ell_L
\end{aligned}
\end{equation} 
We can solve it based on \cite{10885963}, as follows
\begin{equation}
\begin{aligned}
\small
\label{95}
V \leq& V \left(t_0\right) e^{-\rho({t-t_0})}+{\ell} \int_{t_0}^t e^{-{\rho}(t-T)} dT 
\end{aligned}
\end{equation}
Because $e^{-{\rho}(t-T)}$ is always decreasing,
\begin{equation}
\begin{aligned}
\small
\label{96}
{V} \leq&  {V}\left(t_0\right) e^{-\left\{{\rho}\left({t-t_0}\right)\right\}}+ \hspace{0.1cm}{\ell} \hspace{0.1cm} {{\rho}}^{-1}
\end{aligned}
\end{equation}
From \eqref{2dfdds2}, we can say
\begin{equation}
\begin{aligned}
\small
\label{9asdas7}
\hspace{-0.3cm}\sum_{i:R,L} \log \left(\frac{o_i^2}{o^2_i - e^2_i}\right) \leq & 2( {V}\left(t_0\right) e^{-\left\{{\rho}\left({t-t_0}\right)\right\}}+ \hspace{0.1cm}{{\ell}} \hspace{0.1cm} {{\rho}}^{-1})
\end{aligned}
\end{equation}
Thus, based on \textit{Definition 1} and Eq. \eqref{9asdas7}, \textit{Theorem 1} is proved. Note that $\sum_{i=R,L}\log \left(\frac{o_i^2}{o^2_i - e^2_i}\right)$ reaches a defined region $G\left({\tau}\right)$ in uniformly exponential convergence, such that
$G\left({\tau}\right):=\left\{\sum_{i:R,L} \log \left(\frac{o_i^2}{o^2_i - e^2_i}\right) \leq \bar{\tau}_i := {2 \hspace{0.1cm}{{\ell}} \hspace{0.1cm} {{\rho}}^{-1}}\right\}$.

\begin{table*}[t]
    \centering
    \caption{Absolute pose error (APE) in meters for different SLAM methods.}
    \small 
    \renewcommand{\arraystretch}{0.6} 
    \begin{tabular}{c|cccccc}
        \toprule
        \textbf{Sequence} & \textbf{LIO-SAM} & \textbf{FAST-LIO2} & \textbf{ORB-SLAM3} & \textbf{ORB-SLAM3 inertial} & \textbf{VINS-Fusion} & \textbf{VINS-Fusion inertial} \\
        \midrule
        1  & 0.217127  & 0.409905  & \textbf{0.164774}  & 0.485586  & 0.474715  & 0.440113  \\
        2  & \textbf{0.86914}  & 1.754343  & Failed  & 1.715077  & 2.216691  & 1.925925  \\
        3  & 0.81162  & 1.555683  & \textbf{0.542835}  & 1.252041  & 1.387264  & 1.23698  \\
        4  & 0.417588  & 0.70375  & \textbf{0.280474}  & 0.516507  & 0.529805  & 0.454944  \\
        5  & 0.681154  & 1.664873  & \textbf{0.440216}  & 1.520726  & 1.743379  & 1.691131  \\
        6  & 0.869949  & 1.886515  & \textbf{0.735056}  & 1.968528  & 3.277011  & 2.850892  \\
        7  & \textbf{0.088294}  & 0.189954  & 0.103472  & 0.198334  & 1.130662  & 0.889479  \\
        8  & 0.941384  & 0.993397  & \textbf{0.879032}  & 1.466838  & 2.90835  & 2.532334  \\
        9  & \textbf{0.768462}  & 0.82185  & 0.851645  & 1.379856  & 1.742125  & 1.593643  \\
        10 & 0.903748  & 1.524194  & \textbf{0.860641}  & Failed  & 3.012128  & 2.583019  \\
        \midrule
        \textbf{Mean} & \textbf{0.6568466}  & 1.1504464  & \textbf{0.53979389}  & 1.16705478  & 1.842213  & 1.619846  \\
        \bottomrule
    \end{tabular}
    \label{tab:ape_results}
\end{table*}

\begin{figure*} [t]
    \centering
    \includegraphics[width=0.75\textwidth, height=4.3cm]{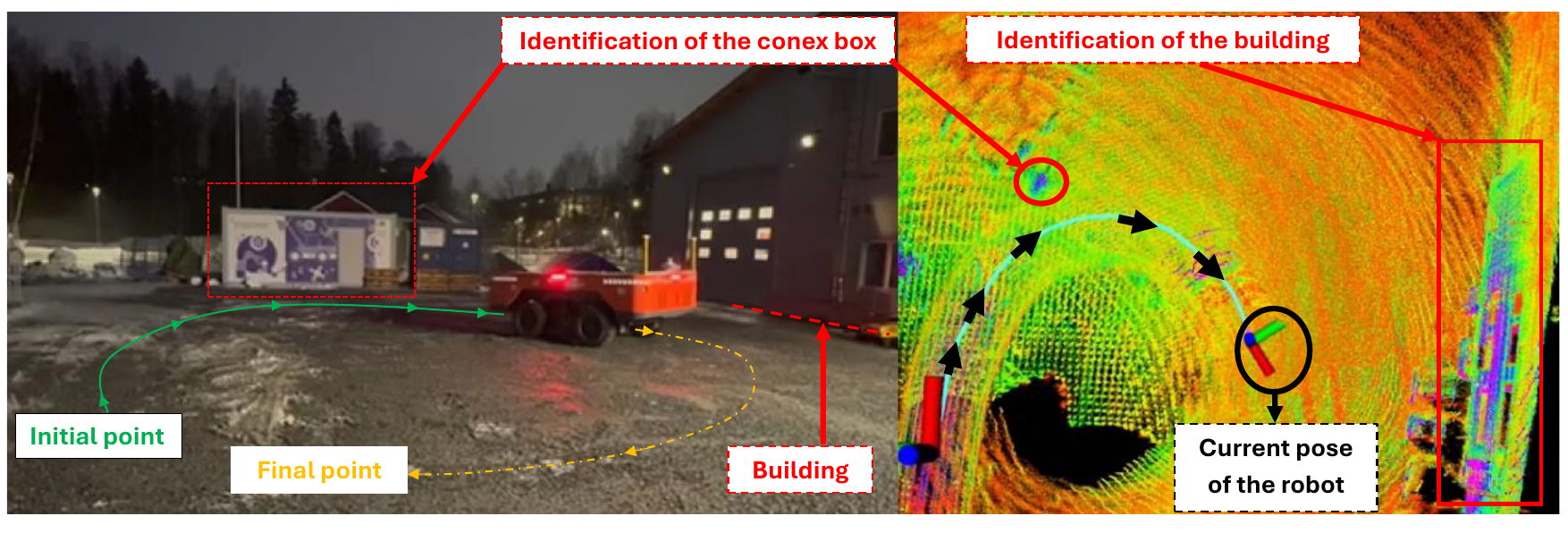}
    \caption{The operation under low-light conditions and on soft terrain using the proposed MRNC framework (see Fig. \ref{schematicwhole}). The relevant video is available at: \texttt{https://youtube.com/shorts/0bQO6-SyoO8?feature=share}.}
    \label{daasff}
\end{figure*}

\section{Experimental Result Analysis}
\begin{figure} [h]
    \centering
    \scalebox{0.95}
    {\includegraphics[trim={0cm 0.0cm 0.0cm
    0cm},clip,width=\columnwidth]{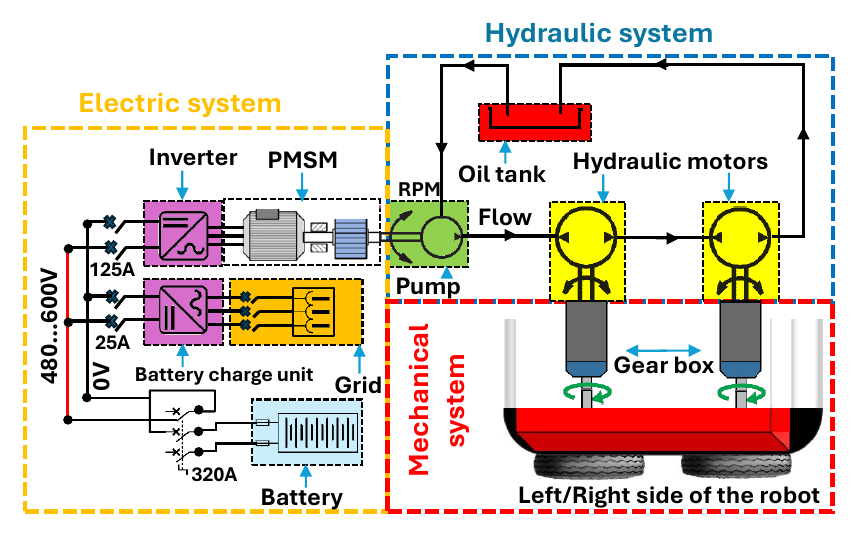}}
    \caption{Mechanical structure of each side of studied MPD.}
    \label{MPD_virtual}
\end{figure}

\begin{figure}[h] 
  \centering
\scalebox{0.65}
    {\includegraphics[trim={0cm 0.0cm 0.0cm
    0cm},clip,width=\columnwidth]{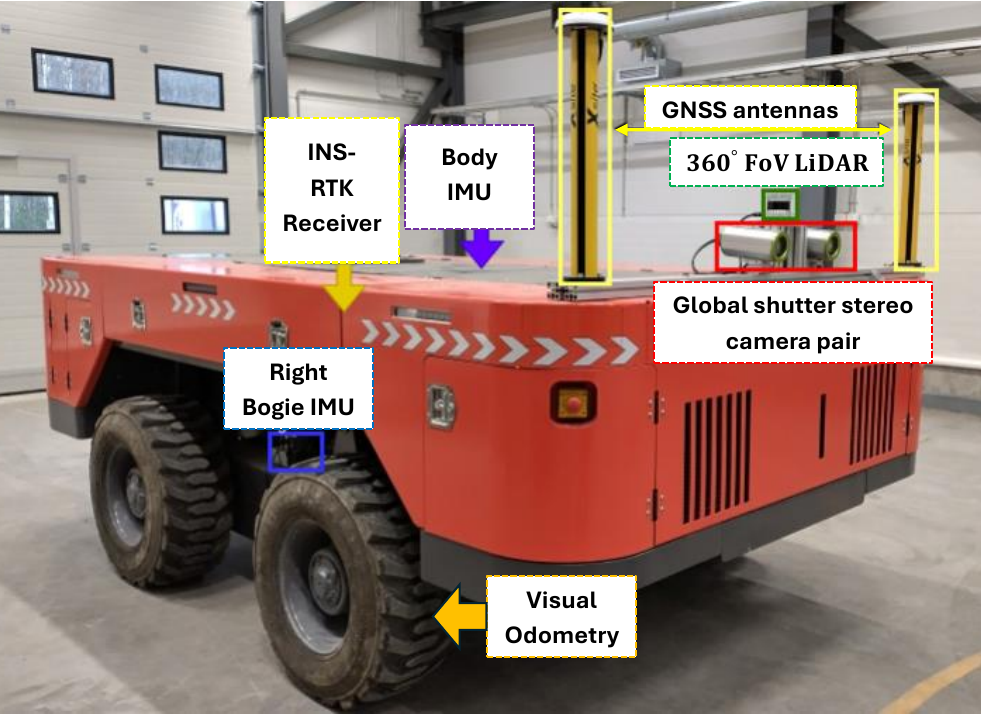}}
  \caption{The dimensions of the MPD are $2 m \times 4 m \times 1.5 \mathrm{~m}$.}
  \label{MPD_virtuall}
\end{figure}

The studied SSWMR, called the multi-purpose deployer (MPD), is a heavy-duty vehicle with a weight of $4,836$ kg and a maximum speed of $0.97 \mathrm{~m} / \mathrm{s}$. The wheels are connected to bogies on each side, which ensures contact with the ground when traversing uneven terrain. The wheels are powered by hydraulic motors. The mechanical structure and sensor setup of the MPD are presented in Figs. \ref{MPD_virtual} and \ref{MPD_virtuall}, respectively. The pump flow model without uncertainties due to leakage, compressibility, or other inefficiencies is $Q_p=V_p n_p$. where $V_p= 1.8 \times 10^{-5} \mathrm{m}^3 / \mathrm{rev}$ is the displacement volume. This represents the fluid volume the pump moves per revolution of its shaft. $n_p$ is the control input and is the rotational speed of the pump shaft, measured in revolutions per minute (RPM).
The required volumetric flow rate for a fixed-displacement motor is specified as follows:
\begin{equation}
\begin{aligned}
\small
\label{equation:adjusted_velocities}
V_p n_p = Q_p=Q_m=\omega_m \frac{V_m}{2 \pi} =\bar{v}_i \frac{\bar{G}}{r} \frac{V_m}{2 \pi} 
\end{aligned}
\end{equation}
where $\bar{G}$ is the gear ratio for conversion between the wheel velocity and the motor velocity. $V_m$ is the motor displacement and $r$ is the radius of the wheel. Thus, from \eqref{2fdgdffg0}, the modeling term of the proposed control is $n_p = f_i = \bar{v}_i \frac{G}{r} \frac{V_m}{2 \pi} \frac{1}{V_p}$, and the control signal is $n^*_p = \mathcal{U}_i = u_i + n_p = u_i + f_i$. Now, from Eq. \eqref{2fdgdffg0}, we design the model-based control with safety constraints, based on machine modeling specifications, as
\begin{equation}
\begin{aligned}
\small
\label{equation:adjusted_velocities}
\mathcal{U}^*_i = \lambda_i [u_i+f_i]+\bar{\lambda}_i = \lambda_i [u_i+ 3000 \hspace{0.1cm} \bar{v}_i] +\bar{\lambda}_i 
\end{aligned}
\end{equation}

The hardware of the perception system is divided into two separate host PCs. The embedded Beckhoff PC reads the IMU data from bogies and the chassis IMUs through the EtherCAT protocol. The measurements are then sent using User Datagram Protocol (UDP) to the Linux side of the system, which is powered by an industrial PC (IPC).
The Ouster OSO-128 LiDAR is connected to the Linux PC through a multi-gigabit switch to ensure enough bandwidth for point clouds. The data from the LiDAR are read utilizing the ROS framework in UDP to transfer the data from sensors to the host machine. 
The internal IMU of the Ouster LiDAR is also included. The external IMUs are read at $1000$ Hz due to their usage in control applications.
The LiDAR IMU is read at 100 Hz by default. 
The IMU for the chassis is ADIS16475-2, configured in its default settings of providing $1024 \times 128$ points at $10$ Hz. Various SLAM methods within the MRNC of the studied SSWMR are evaluated across 10 sequences in daylight conditions. The comparative APE is presented in Table \ref{tab:ape_results}, where the last four algorithms use stereo cameras instead of LiDar. Although the lowest mean APE is attained by ORB-SLAM3 in stereo mode and Lidar-inertial LIO-SAM, using ORB-SLAM3 in low-light conditions is challenging.
Eight waypoints, depending on the initial pose, and a $1$ m lookahead distance are chosen for the pure pursuit control to achieve a circular trajectory with about $5$ m radius. In addition, for safety reasons, the maximum linear velocity and the desired velocity are set to $0.4$ m/s and $0.38$ m/s (about $50 \%$ of the robot's capability), respectively. The safely predefined control performance in terms of over-shooting and steady-state error is: $o_i = 0.19 e^{-0.00009} + 0.11$. The actuator control effort constraint is $|\mathcal{U}^*_i| < 1250$ RPM. 
The control gains are $\beta_i=1.2$ and $\kappa_i = 5.2$, $N_i = 9$, $\alpha_{i} = [2 \hspace{0.1cm} {rand}-1]$, and $\gamma_{i}=0.13$, where $rand$ outputs a variable between $0$ and $1$.

\begin{figure}[h] 
  \centering
\scalebox{0.9}
    {\includegraphics[trim={0cm 0.0cm 0.0cm
    0cm},clip,width=\columnwidth]{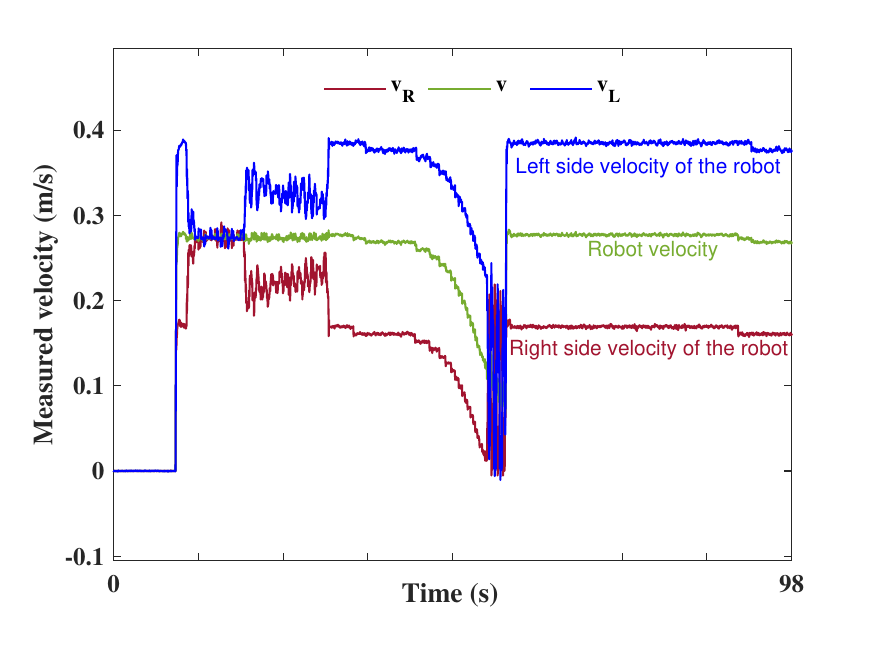}}
  \caption{Measured velocities through wheel odometry sensors.}
  \label{vellll}
\end{figure}

\begin{figure}[h] 
  \centering
\scalebox{0.9}
    {\includegraphics[trim={0cm 0.0cm 0.0cm
    0cm},clip,width=\columnwidth]{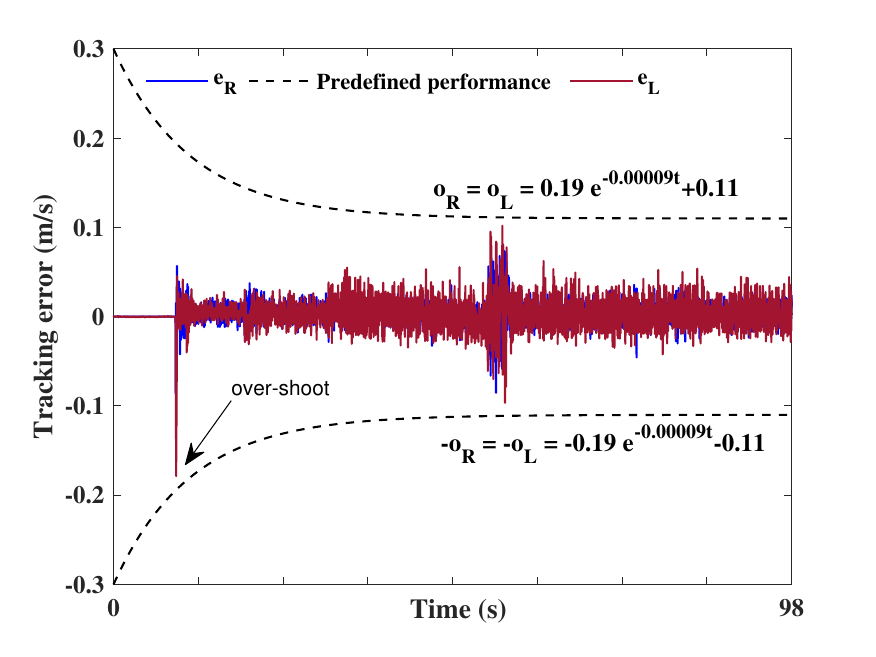}}
  \caption{RAID tracking control performance under predefined performance constraints.}
  \label{errdfan}
\end{figure}

Fig. \ref{daasff} shows the entire operation as well as the recorded data from the employed LiDar-inertial SLAM algorithm. Note that it successfully identifies obstacles that can be used for designing collision-avoidance algorithms in future work. The conditions involves low light, and the terrain is wet and soft.
The recorded velocities are indicated in Fig. \ref{vellll} in the autonomous operation. As defined in the pure pursuit algorithm, the robot has the right side heading on the circle trajectory. Thus, the linear velocity of the left side is greater than that of the right side. The actual velocities at approximately 10-20s and again at 50s show wheel slippage due to the wet and soft ground; however, the RIAD control maintains stable tracking despite these disturbances.
Fig. \ref{errdfan} indicates the control tracking error between the measured wheel velocity (obtained from the wheel odometry sensor) and the desired velocity for each side (generated by the pure pursuit and inverse kinematics algorithms). The results show that both sides adhere to the safely predefined control performance.

\begin{figure}[h] 
  \centering
\scalebox{0.9}
    {\includegraphics[trim={0cm 0.0cm 0.0cm
    0cm},clip,width=\columnwidth]{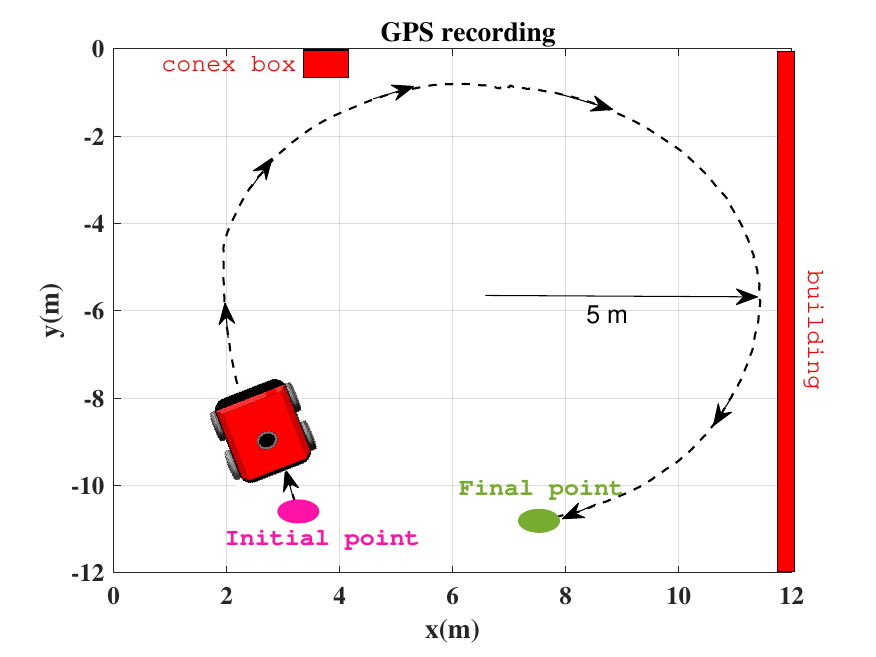}}
  \caption{The passed trajectory in 2D plane recorded by GPS.}
  \label{gpss}
\end{figure}

\begin{figure}[h] 
  \centering
\scalebox{0.86}
    {\includegraphics[trim={0cm 0.0cm 0.0cm
    0cm},clip,width=\columnwidth]{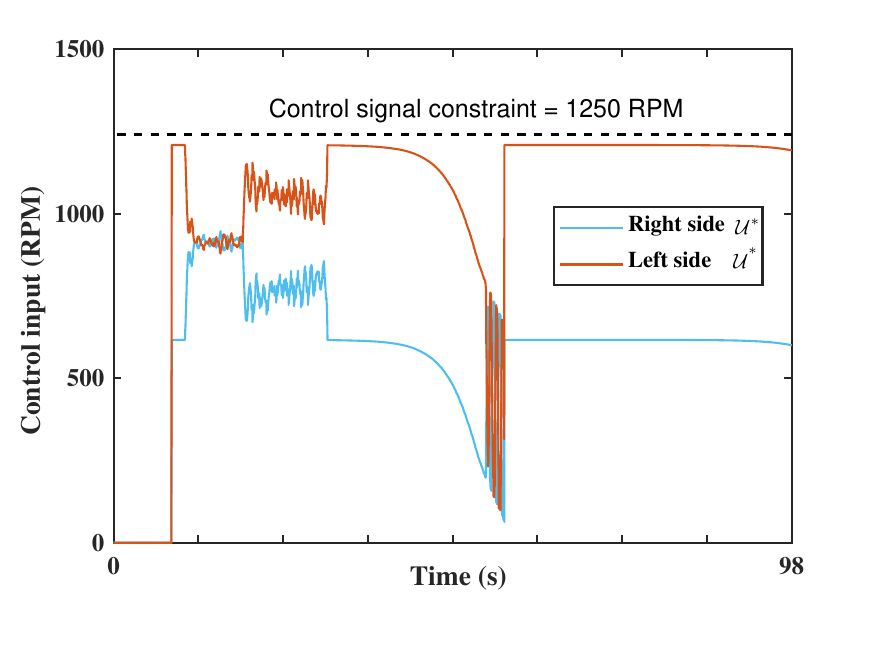}}
  \caption{RAID control signals for each side of the SSWMR.}
  \label{sadasdsa}
\end{figure}

Fig. \ref{gpss} shows the trajectory of the robot in a 2D plane, recorded by GPS, forming a roughly circular shape with a radius of approximately 5 m. Wheel slippage from the location (2,-8) to (2, -4) and at (7, -1) are evident, while the RIAD maintains tracking stability under wheel slippage. The generated output efforts of the RAID control for each side of the pumps based on RPM are shown in Fig. \ref{sadasdsa}, maintaining the signals within the safety actuator constraint of $1250$ RPM. The performance of the RAID control in the same condition is compared with three control algorithms: an adaptive predictive control (APC) provided in \cite{zhu2024backstepping}, PID control, and a fractional-order prescribed-time controller (FOPTC) provided in \cite{ge2023prescribed}. The comparative results are summarized in Table \ref{adasdsda}, validating the proposed RAID control efficacy.

\begin{table}[ht]
\small
\centering
\caption{Control performance metrics in the experiment.}
    \renewcommand{\arraystretch}{0.7} 
    \begin{tabular}{c|cccc}
\hline
\textbf{Stability Criteria} & RAID & APC & PID & FOPTC\\
\hline
\textbf{Settling time} & $3.2$ s & $3.6$ s & $5.8$ s & $5.3$ s\\
\textbf{Overshoot}  & $1.7 \%$ & $2.1\%$ &  $2.3\%$ & $1.8\%$\\
\textbf{Steady-state error} & $0.043$& $0.069$  & $0.12$& $0.056$ \\
\hline
\end{tabular}
\label{adasdsda}
\end{table}

\section{Conclusion}
To meet rigorous safety standards and the complexity of SSWMRs, this paper proposes a novel integrated MRNC framework, in which all components are actively engaged in real-time execution, including: 1) a LiDAR-inertial SLAM algorithm, 2) an effective path-following control system, 3) inverse kinematics for transferring linear and angular velocity commands, and 4) a novel robust RAID control system to enforce in-wheel actuation systems. The proposed RAID control within the MRNC framework constrained the AI-generated tracking performance within predefined overshoot and steady-state error boundaries while ensuring robustness and system stability by compensating for uncertainties and external forces.
\bibliographystyle{IEEEtran}
\bibliography{ref}

\begin{thebibliography}{10}
\providecommand{\url}[1]{#1}
\csname url@samestyle\endcsname
\providecommand{\newblock}{\relax}
\providecommand{\bibinfo}[2]{#2}
\providecommand{\BIBentrySTDinterwordspacing}{\spaceskip=0pt\relax}
\providecommand{\BIBentryALTinterwordstretchfactor}{4}
\providecommand{\BIBentryALTinterwordspacing}{\spaceskip=\fontdimen2\font plus
\BIBentryALTinterwordstretchfactor\fontdimen3\font minus \fontdimen4\font\relax}
\providecommand{\BIBforeignlanguage}[2]{{%
\expandafter\ifx\csname l@#1\endcsname\relax
\typeout{** WARNING: IEEEtran.bst: No hyphenation pattern has been}%
\typeout{** loaded for the language `#1'. Using the pattern for}%
\typeout{** the default language instead.}%
\else
\language=\csname l@#1\endcsname
\fi
#2}}
\providecommand{\BIBdecl}{\relax}
\BIBdecl

\bibitem{khan2021comprehensive}
R.~Khan, F.~M. Malik, A.~Raza, and N.~Mazhar, ``Comprehensive study of skid-steer wheeled mobile robots: development and challenges,'' \emph{Industrial Robot: The International Journal of Robotics Research and Application}, vol.~48, no.~1, pp. 142--156, 2021.

\bibitem{labbe2019rtab}
M.~Labb{\'e} and F.~Michaud, ``Rtab-map as an open-source lidar and visual simultaneous localization and mapping library for large-scale and long-term online operation,'' \emph{Journal of Field Robotics}, vol.~36, no.~2, pp. 416--446, 2019.

\bibitem{mohamed2019survey}
S.~A. Mohamed, M.-H. Haghbayan, T.~Westerlund, J.~Heikkonen, H.~Tenhunen, and J.~Plosila, ``A survey on odometry for autonomous navigation systems,'' \emph{IEEE Access}, vol.~7, pp. 97\,466--97\,486, 2019.

\bibitem{cadena2016past}
C.~Cadena, L.~Carlone, H.~Carrillo, Y.~Latif, D.~Scaramuzza, J.~Neira, I.~Reid, and J.~J. Leonard, ``Past, present, and future of simultaneous localization and mapping: Toward the robust-perception age,'' \emph{IEEE Transactions on Robotics}, vol.~32, no.~6, pp. 1309--1332, 2016.

\bibitem{khattak2020complementary}
S.~Khattak, H.~Nguyen, F.~Mascarich, T.~Dang, and K.~Alexis, ``Complementary multi--modal sensor fusion for resilient robot pose estimation in subterranean environments,'' in \emph{2020 International Conference on Unmanned Aircraft Systems (ICUAS)}.\hskip 1em plus 0.5em minus 0.4em\relax IEEE, 2020, pp. 1024--1029.

\bibitem{shan2020lio}
T.~Shan, B.~Englot, D.~Meyers, W.~Wang, C.~Ratti, and D.~Rus, ``Lio-sam: Tightly-coupled lidar inertial odometry via smoothing and mapping,'' in \emph{2020 IEEE/RSJ International Conference on Intelligent Robots and Systems (IROS)}.\hskip 1em plus 0.5em minus 0.4em\relax IEEE, 2020, pp. 5135--5142.

\bibitem{huskic2019gerona}
G.~Huski{\'c}, S.~Buck, and A.~Zell, ``Gerona: generic robot navigation: a modular framework for robot navigation and control,'' \emph{Journal of Intelligent \& Robotic Systems}, vol.~95, no.~2, pp. 419--442, 2019.

\bibitem{shahna2025fault}
M.~H. Shahna, P.~Mustalahti, and J.~Mattila, ``Fault-tolerant control for system availability and continuous operation in heavy-duty wheeled mobile robots,'' \emph{arXiv preprint arXiv:2502.03278}, 2025.

\bibitem{lee2024learning}
J.~Lee, M.~Bjelonic, A.~Reske, L.~Wellhausen, T.~Miki, and M.~Hutter, ``Learning robust autonomous navigation and locomotion for wheeled-legged robots,'' \emph{Science Robotics}, vol.~9, no.~89, p. eadi9641, 2024.

\bibitem{shahna2024integrating}
M.~H. Shahna, S.~A.~A. Kolagar, and J.~Mattila, ``Integrating deeprl with robust low-level control in robotic manipulators for non-repetitive reaching tasks,'' in \emph{2024 IEEE International Conference on Mechatronics and Automation (ICMA)}.\hskip 1em plus 0.5em minus 0.4em\relax IEEE, 2024, pp. 329--336.

\bibitem{carabantes2020black}
M.~Carabantes, ``Black-box artificial intelligence: an epistemological and critical analysis,'' \emph{AI \& Society}, vol.~35, no.~2, pp. 309--317, 2020.

\bibitem{ISO81283}
{International Organization for Standardization}, ``{ISO/IEC TR} 5469 - {A}rtificial intelligence — {F}unctional safety and {AI} systems,'' 2024.

\bibitem{shi2024barrier}
Q.~Shi, R.~Wang, X.~Li, and J.~Yang, ``Barrier lyapunov function-based adaptive optimized control for full-state and input-constrained dynamic positioning of marine vessels with simulation and model-scale tests,'' \emph{Ocean Engineering}, vol. 301, p. 117534, 2024.

\bibitem{han2013barrier}
S.~I. Han, J.~Y. Cheong, and J.~M. Lee, ``Barrier lyapunov function-based sliding mode control for guaranteed tracking performance of robot manipulator,'' \emph{Mathematical Problems in Engineering}, vol. 2013, no.~1, p. 978241, 2013.

\bibitem{shahna2025model}
M.~H. Shahna, J.-P. Humaloja, and J.~Mattila, ``Model reference-based control with guaranteed predefined performance for uncertain strict-feedback systems,'' \emph{arXiv preprint arXiv:2502.03263}, 2025.

\bibitem{shahna2024robusttttsdf}
M.~H. Shahna, P.~Mustalahti, and J.~Mattila, ``Robust sensor-limited control with safe input-output constraints for hydraulic in-wheel motor drive mobility systems,'' \emph{arXiv preprint arXiv:2409.11823}, 2024.

\bibitem{zhang2019low}
J.-X. Zhang and G.-H. Yang, ``Low-complexity tracking control of strict-feedback systems with unknown control directions,'' \emph{IEEE Transactions on Automatic Control}, vol.~64, no.~12, pp. 5175--5182, 2019.

\bibitem{zhang2017low}
J.~Zhang and S.~Singh, ``Low-drift and real-time lidar odometry and mapping,'' \emph{Autonomous Robots}, vol.~41, pp. 401--416, 2017.

\bibitem{grisetti2010tutorial}
G.~Grisetti, R.~K{\"u}mmerle, C.~Stachniss, and W.~Burgard, ``A tutorial on graph-based slam,'' \emph{IEEE Intelligent Transportation Systems Magazine}, vol.~2, no.~4, pp. 31--43, 2010.

\bibitem{krecht2020possible}
R.~Krecht, C.~Hajdu, and {\'A}.~Ballagi, ``Possible control methods for skid-steer mobile robot platforms,'' in \emph{2020 2nd IEEE International Conference on Gridding and Polytope Based Modelling and Control (GPMC)}.\hskip 1em plus 0.5em minus 0.4em\relax IEEE, 2020, pp. 31--34.

\bibitem{coulter1992implementation}
R.~C. Coulter, ``Implementation of the pure pursuit path tracking algorithm,'' 1992.

\bibitem{huang2019practical}
Y.~Huang and Y.~Liu, ``Practical tracking via adaptive event-triggered feedback for uncertain nonlinear systems,'' \emph{IEEE Transactions on Automatic Control}, vol.~64, no.~9, pp. 3920--3927, January 2019.

\bibitem{corless1993bounded}
M.~Corless and G.~Leitmann, ``Bounded controllers for robust exponential convergence,'' \emph{Journal of Optimization Theory and Applications}, vol.~76, no.~1, pp. 1--12, 1993.

\bibitem{shahna2024robustness}
M.~H. Shahna, M.~Bahari, and J.~Mattila, ``Robustness-guaranteed observer-based control strategy with modularity for cleantech emla-driven heavy-duty robotic manipulator,'' \emph{IEEE Transactions on Automation Science and Engineering}, 2024.

\bibitem{heydari2024robusttt}
M.~Heydari~Shahna, M.~Bahari, and J.~Mattila, ``Robust decomposed system control for an electro-mechanical linear actuator mechanism under input constraints,'' \emph{International Journal of Robust Nonlinear Control}, 2024.

\bibitem{ren2010adaptive}
B.~Ren, S.~S. Ge, K.~P. Tee, and T.~H. Lee, ``Adaptive neural control for output feedback nonlinear systems using a barrier lyapunov function,'' \emph{IEEE Transactions on Neural Networks}, vol.~21, no.~8, pp. 1339--1345, July 2010.

\bibitem{shahna2024robust}
M.~H. Shahna, P.~Mustalahti, and J.~Mattila, ``Robust model-free control framework with safety constraints for a fully electric linear actuator system,'' in \emph{2024 IEEE 21st International Power Electronics and Motion Control Conference (PEMC)}.\hskip 1em plus 0.5em minus 0.4em\relax IEEE, 2024, pp. 1--10.

\bibitem{10885963}
M.~H. Shahna and J.~Mattila, ``Exponential auto-tuning fault-tolerant control of n degrees-of-freedom manipulators subject to torque constraints,'' in \emph{2024 IEEE 63rd Conference on Decision and Control (CDC)}, 2024, pp. 7263--7270.

\bibitem{zhu2024backstepping}
Z.~Zhu and Z.~Zhu, ``Backstepping model-free adaptive control for a class of second-order nonlinear systems,'' \emph{International Journal of Robust and Nonlinear Control}, vol.~34, no.~5, pp. 3057--3073, 2024.

\bibitem{ge2023prescribed}
M.~Ge, H.~Xu, and Q.~Song, ``Prescribed-time control of four-wheel independently driven skid-steering mobile robots with prescribed performance,'' \emph{Nonlinear Dynamics}, vol. 111, no.~22, pp. 20\,991--21\,005, 2023.

\end{thebibliography}

\end{document}